\documentclass[letterpaper, 10 pt, conference]{ieeeconf}  

\IEEEoverridecommandlockouts                              

\makeatletter
\let\NAT@parse\undefined
\makeatother
\usepackage{graphics} 
\usepackage[font={footnotesize},labelfont={bf,footnotesize},figurename=Fig.,tablename=Table]{caption}
\usepackage{graphicx}
\usepackage{grffile} 
\usepackage{amsmath} 
\usepackage{amssymb}  
\usepackage{color}
\usepackage{mathtools} 
\usepackage{booktabs,ragged2e}
\usepackage{breqn}
\usepackage{textgreek}
\usepackage{arydshln}
\usepackage{tabularx,booktabs}
\usepackage{graphics}
\usepackage{amsmath}
\usepackage{gensymb}
\usepackage[normalem]{ulem} 
\usepackage{pifont}
\usepackage{booktabs}
\usepackage{multirow}

\usepackage{lineno}
\usepackage{cite}
\usepackage{bm}
\usepackage{url}

\usepackage[noend]{algpseudocode}
\usepackage{algorithm}
\usepackage{makecell}
\usepackage[square, comma, numbers,sort&compress]{natbib}
\usepackage{multirow}
\let\labelindent\relax
\usepackage{enumitem}
\usepackage[table,dvipsnames]{xcolor}
\usepackage[pagebackref=true,breaklinks=true,colorlinks,bookmarks=false]{hyperref}
\usepackage{comment}
\usepackage{svg}
\hypersetup{
linkcolor=BrickRed
,citecolor=Green
,filecolor=Mulberry
,urlcolor=NavyBlue
,menucolor=BrickRed
,runcolor=Mulberry
,linkbordercolor=BrickRed
,citebordercolor=Green
,filebordercolor=Mulberry
,urlbordercolor=NavyBlue
,menubordercolor=BrickRed
,runbordercolor=Mulberry
}

\usepackage[]{todonotes}

\usepackage{marginnote}

\usepackage[normalem]{ulem} 
\newcommand{\algname}{SWIFT}

\usepackage{xcolor}
\usepackage{soul}
\usepackage{balance}
\usepackage{bbm}
\usepackage{dsfont}

\title{\LARGE \bf
   Soft Robotic Dynamic In-Hand Pen Spinning 
}
\author{\small Yunchao Yao\textsuperscript{1 \ding{41}} \quad \small Uksang Yoo \textsuperscript{1} \quad 
Jean Oh \textsuperscript{1} \quad
  Christopher G Atkeson \textsuperscript{1} \quad Jeffrey Ichnowski \textsuperscript{1}
\thanks{ \textsuperscript{\ding{41}}Corresponding author.}%
\thanks{$^{1}$Robotics Institute, Carnegie Mellon University, Pittsburgh, USA
{\tt\footnotesize \{yunchaoy,uyoo,hyaejino,cga,jichnows\}@andrew.cmu.edu}}
}

\begin{document}

\maketitle
\thispagestyle{empty}
\pagestyle{empty}
\renewcommand{\baselinestretch}{0.979} 

\begin{abstract}
Dynamic in-hand manipulation remains a challenging task for soft robotic systems that have demonstrated advantages in safe compliant interactions but struggle with high-speed dynamic tasks. In this work, we present SWIFT, a system for learning dynamic tasks using a soft and compliant
robotic hand. Unlike previous works that rely on simulation, quasi-static actions and precise object models, the proposed system learns to spin a pen through trial-and-error using only real-world data without requiring explicit prior knowledge of the pen's physical attributes. With self-labeled trials sampled from the real world, the system discovers the set of pen grasping and spinning primitive parameters that enables a soft hand to spin a pen robustly and reliably. After 130 sampled actions per object, \algname{} achieves 100\,\% success rate across three pens with different weights and weight distributions, demonstrating the system's generalizability and robustness to changes in object properties. The results highlight the potential for soft robotic end-effectors to perform dynamic tasks including rapid in-hand manipulation.  We also demonstrate that \algname{} generalizes to spinning items with different shapes and weights such as a brush and a screwdriver which we spin with 10/10 and 5/10 success rates respectively. Videos, data, and code are available at  \url{https://soft-spin.github.io}.

\end{abstract}

\renewcommand{\baselinestretch}{0.979} 
\section{Introduction}

In-hand dexterity is crucial for many tasks common in our daily lives~\cite{nakamura2017complexities} and the ability to re-orient objects in the hand and re-grasp them is useful to perform these tasks efficiently and effectively~\cite{chen2022system,sundaralingam2018geometric}.
The compliance of soft robotic end-effectors' deformable fingers allows them to be robust to perturbations~\cite{bhatt2022surprisingly, sieler2023dexterous} and enables them to interact with their environments safely~\cite{sinatra2019ultragentle, yoo2024moe}. However, compliance makes it hard to move the fingers both quickly and accurately. Previous works on soft robotic end-effector dexterity have focused on largely slow quasi-static tasks such as grasping and slowly reorienting objects~\cite{bhatt2022surprisingly, homberg2019robust}. Such limitations underscore the gap between soft robotic end-effectors and human hands that can fully exploit the dynamics of objects to efficiently re-orient and re-grasp various tools and objects. 

\begin{figure}
    \centering
    \includegraphics[width=1.0\linewidth]{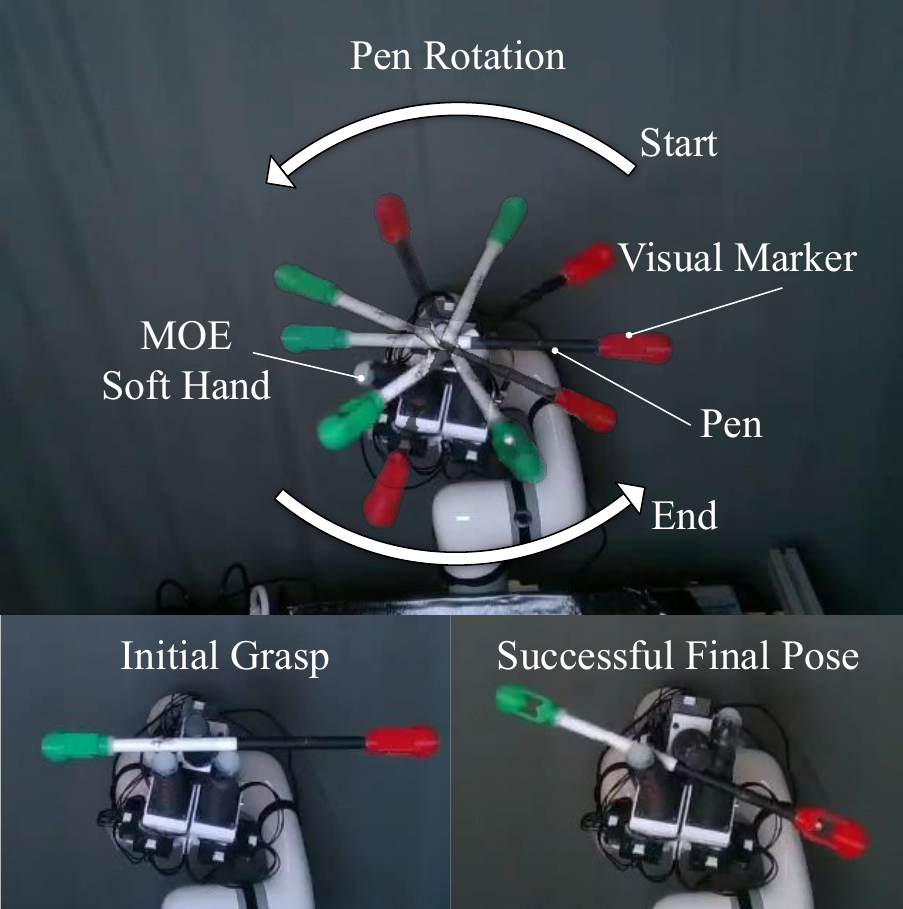}
    \caption{\algname{} tackles the problem of high-speed dynamic in-hand partially non-prehensile manipulation with soft robotic hands. Using a soft multi-finger gripper, the robot grasps a pen. Then, using a learned action sequence, rapidly rotates the pen around a finger and catches it. 
    } 
    \label{fig:front}
    \vspace{-0.5cm}
\end{figure}

Pen spinning is a challenging dynamic task that humans often have difficulty mastering. As a case study of how to enable soft robots to perform fast dynamic tasks (Fig~\ref{fig:front}), it can suggest methods for tackling fast manipulation tasks with a soft manipulator. Approaching the problem of in-hand object re-orientation dynamically allows the robot to perform the task efficiently in one continuous action sequence as demonstrated in other dynamic manipulation tasks~\cite{bombile2022dual, ha2022flingbot}.
Previous works on exploiting object dynamics for in-hand reorientation of objects relied on knowing the object properties such as its weight and world parameters~\cite{nakatani2023dynamic}. However, in practice, we may not know such parameters a priori. For example, in the case of pen spinning, visual observation may be inadequate to estimate the distribution of the weight and its center of mass to find the appropriate action parameters to successfully spin the pen. Moreover, the spinning motions usually last for less than a second, requiring high-speed sensing that are not always available and making close-loop control impractical. 
Rather than relying on knowing these parameters prior to the interactions or close loop contorl, we present Soft-hand With In-hand Fast re-orienTation (SWIFT), a system that learns where to grasp and how to dynamically manipulate objects autonomously through trial and error. A crucial component of \algname{} is the softness and compliance of the soft Multi-finger Omnidirectional End-effector (MOE) \cite{yoo2024moe} 
enabling the system's ability to safely interact with the environment to pick up an object reliably and attempt various dynamic manipulations. 

The contact-rich nature of object re-orientation tasks such as pen spinning leads to a sizable gap between simulated environments and the real world, requiring extensive fine-tuning of simulators with real-world data~\cite{wanglessons}. Additionally, the soft robotic end-effector introduces unresolved challenges in not only simulating contact interactions realistically but also in simulating complex soft-body mechanical phenomena such as soft material hysteresis and creep~\cite{liu2021influence}.  
Therefore, we learn pen grasping and dynamic spinning skills with only real-world interactions. To this end, we define the task and desired behavior with an objective function evaluated with RGB-Depth camera feedback. We implement primitives for soft robotic pen spinning designed to reduce the dimension of the search space for a successful pen spin to 8 parameters. We use a soft and compliant robotic hand and a 6 degree-of-freedom robotic arm to safely interact with the environment to repeatedly grasp and attempt to spin the pen. We deploy a evolution-based optimization system to efficiently explore the primitives' parameter space and narrow it down to a locally optimal set of parameters to successfully spin the pen. In experiments, we demonstrate that using the proposed \algname{} system, we can learn to grasp and spin pens even if the properties of the pen such as its weight or weight distribution are different.

To summarize, we make the following contributions in this work:
\begin{enumerate}
    \item Demonstrating a dynamic task of grasping and dynamically spinning a pen in a soft robotic hand,
    \item Developing a self-supervised autonomous process to rapidly learn to spin a pen dynamically with a soft compliant hand, 
    \item Evaluating our approach under a variety of conditions.
\end{enumerate}
\renewcommand{\baselinestretch}{0.979} 
\section{Related Work}

\subsection{Dynamic Manipulation}
Researchers have proposed various tasks and methods for dynamic manipulation of objects such as throwing objects~\cite{liu2024tube, zeng2020tossingbot,bianchi2023softoss}, fast transport of grasped objects~\cite{ichnowski2022gomp} and flinging rope or cloth~\cite{chen2022efficiently, ha2022flingbot,zhang2021robots, chi2024iterative, yamakawa2013dynamic}. In most of these previous works, the robotic arms provided an impulse to the object to move them dynamically~\cite{ichnowski2022gomp}. 

\subsection{Robotic Pen Spinning}

To study in-hand dexterity and dynamic manipulation, recent works propose studying the task of pen spinning~\cite{ishihara2006dynamic,nakatani2023dynamic,wanglessons}. Pen spinning is interesting because it involves many challenging aspects of in-hand dexterity such as contact-rich interactions, partially non-prehensile manipulation, and object dynamics. The researchers generally approached the task in one of two ways: analytical dynamics model-based control~\cite{nakatani2023dynamic} and reinforcement learning aided by simulation environments~\cite{wanglessons}. 
\cite{nakatani2023dynamic} used a rigid robot hand, which is easier to model than the soft hand used in this work.
Reinforcement learning-based approaches allow the researchers to define the task with reward functions that are  hand-crafted~\cite{wanglessons} or semantically produced with language models~\cite{ma2023eureka}. Because of complex contact interactions and object dynamics of pen spinning, learning approaches have struggled with the sim-to-real transfer of policies trained in simulation and have only demonstrated quasi-static pen spinning with slow incremental re-orientation. 

\subsection{Soft Robotic Manipulation}

Researchers have demonstrated the advantages of soft robotic manipulators in various quasi-static tasks such as object grasping~\cite{puhlmann2022rbo} and slow re-orientation of regular objects such as cubes~\cite{bhatt2022surprisingly}. Recent works have demonstrated methods to exploit soft robotic manipulators' inertial dynamics to accomplish tasks such as throwing efficiently~\cite{haggerty2023control, bianchi2024softsling}. Similarly to works in rigid dynamic manipulation, the focus has been on high-velocity control of soft robotic arms~\cite{bruder2019nonlinear}. To our knowledge, this work is the first to explore using soft robotic end-effectors for dynamic in-hand tasks such as pen spinning.

\renewcommand{\baselinestretch}{0.979}

\section{Problem Statement}

The problem is to enable a soft robot hand to perform a high-speed in-hand  rotation of a held cylindrical object.
Specifically, the hand should perform a pen spinning task, similar to the ``Thumbaround'' trick done by humans, where the pen is pushed by the middle finger and spins around the thumb before being caught by the index finger.
We assume that 
the object is
long and cylindrically symmetric with a well-defined major axis and the mass and size is within the hand's grasping and manipulation capabilities.
While an attached manipulator arm positions the hand for repeatable grasps, it does not participate in the high-speed manipulation. We define success as a full rotation of the object without dropping.




\renewcommand{\baselinestretch}{0.979} 
\section{Methods}

\subsection{Soft Hand}
\begin{figure}[t]
  \centering
    \includegraphics[width=1.0\linewidth]{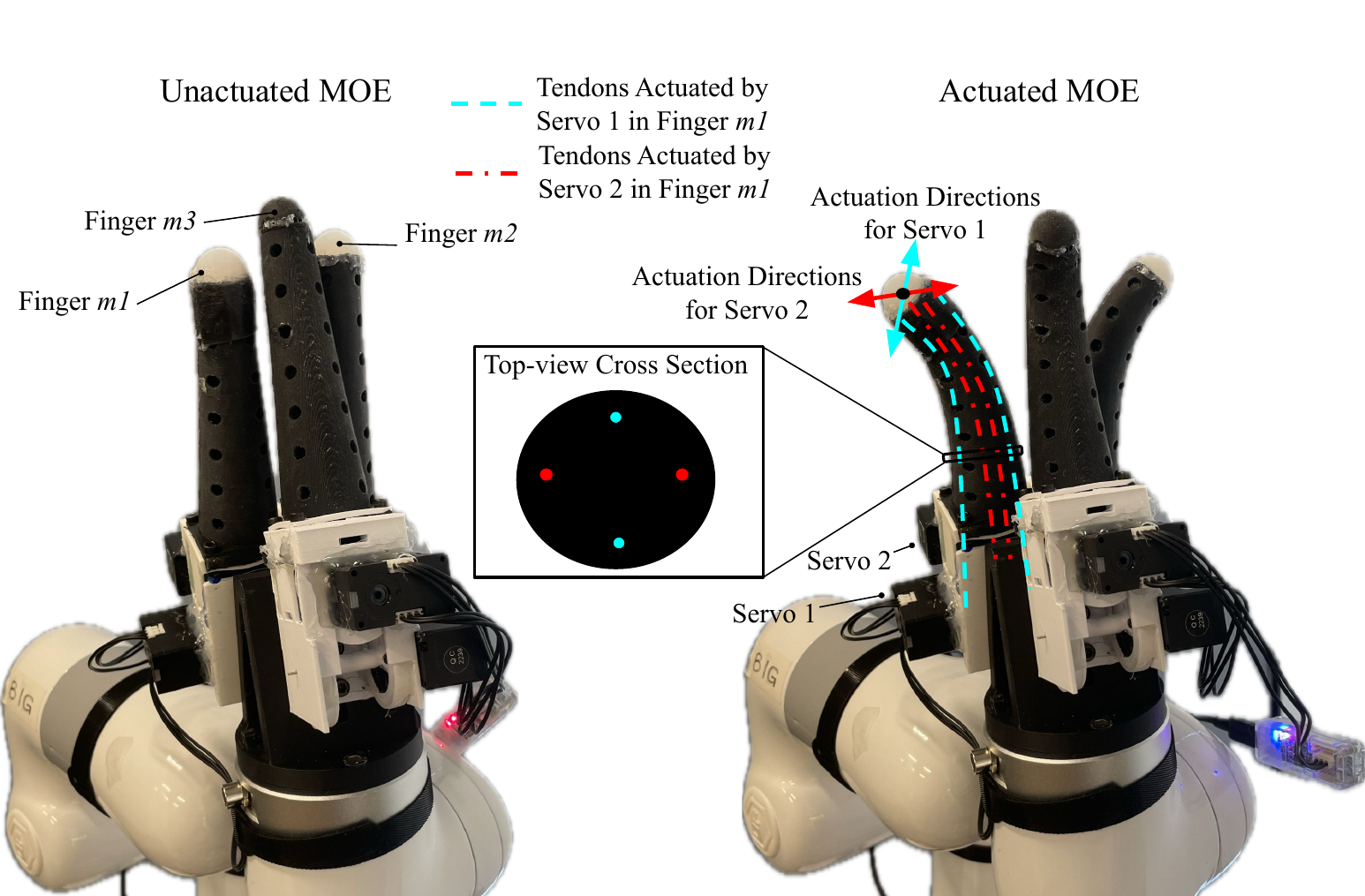}
    \caption{\textbf{Multi-finger Omnidirectional End-effector (MOE)}. The soft hand we used is a three-finger variant of the MOE. Each finger has four tendons actuated by two servo motors, each motor controlling the finger in perpendicular directions.  }
    \label{fig:moe}
\end{figure}

To tackle the problem of 
in-hand dynamic pen spinning we constructed a soft robot hand using the \emph{multi-finger omni-directional end-effector (MOE)}~\cite{yoo2024poe}. The sot robot hand consists of three tendon-driven soft robot fingers, each driven by two servos controlling four tendons. Fig.~\ref{fig:moe} shows the MOE unactuated and actuated by the tendons. The servos pull the tendons to bend the finger in perpendicular planes, and combining the servo motions can actuate each finger tip of MOE hand to reach locations on its semi-hemisphere workspace. 
Two fingers, denoted by $m1$, $m2$, are attached on one side of the hand base, and the last finger, denoted by $m3$, is attached to the other side. We attached the MOE hand to a 6-DOF robot arm.
Fig~\ref{fig:moe} illustrates more details of our MOE and the finger configurations. 

\subsection{Setup and Reset Procedure}
\begin{figure*}[t]
  \centering
     \includegraphics[width=0.99\linewidth]{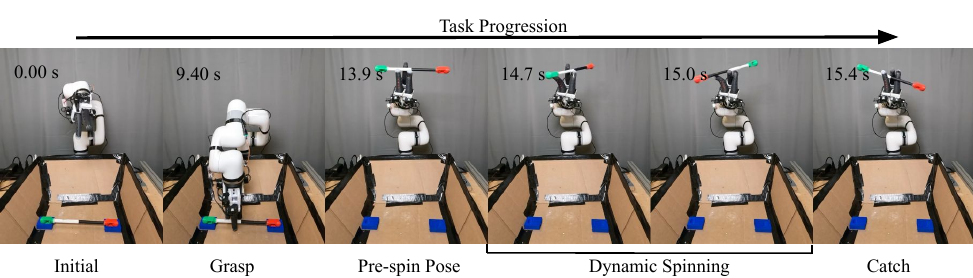}
     \caption{\textbf{Task progression over time}. There are three main stages for each pen-spinning trajectory. We place the pen according to the blue slots fixed on the table, and the robot moves to grasp and move the pen to reach the pre-spin pose with $g$ or pre-defined constant. The MOE fingers then execute $s$ to attempt to spin the pen, and finger $m1$ waits for $d$ seconds before closing to catch the pen. Finally, the robot arm moves to the initial joint configuration, dropping the pen and restarting the cycle. 
     }
    \label{fig:task}
\end{figure*}
\begin{figure}[t]
    \centering
    \includegraphics[width=1.0\linewidth]{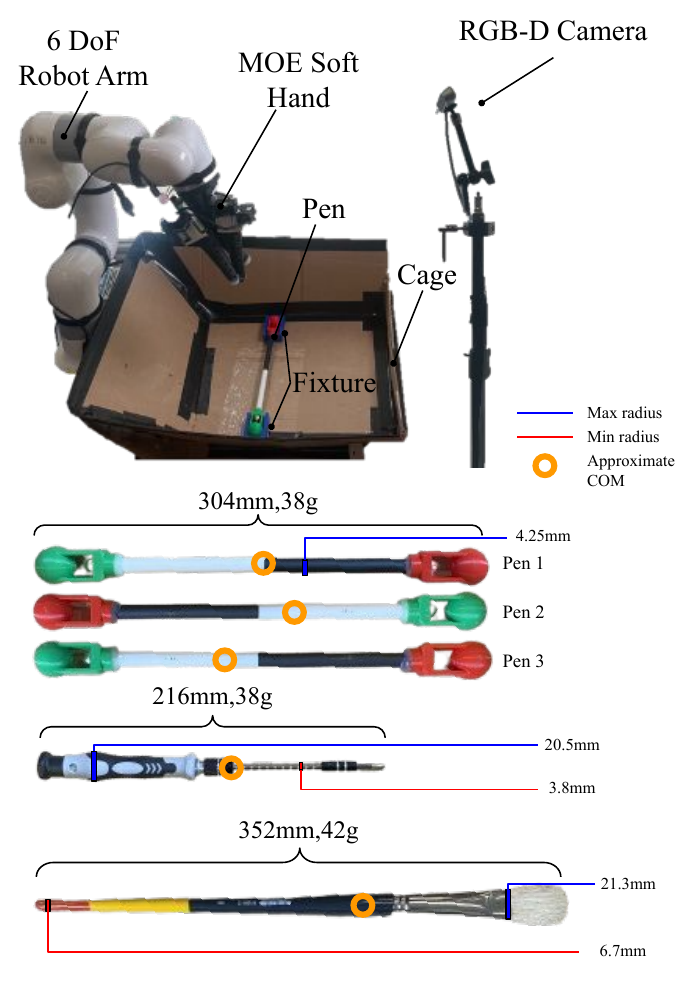}
    \caption{\textbf{Our setup for pen spinning.} Top: A 3-finger MOE soft robotic hand is attached to a 6 degree-of-freedom robot arm to develop a system that can safely interact with the pen and learn to spin it. An RGB-D camera is used to evaluate the performance of the sampled action based on the objective function. 
    The box catches the pen when it is dropped to simplify resetting the system for the next trial. Bottom: the length, radius, weight, and approximate center of mass of each object used in the experiment}
    \label{fig:setup}
\end{figure}

  Before each attempt to spin the pen, we first manually place the pen in a fixed slot on the table (Fig~\ref{fig:task} Initial). This fixturing process results in repeatable grasps. The robot arm then executes a fixed set of movements to move the MOE hand to approximately the center of the pen. The MOE fingers then close to grasp the pen, and the robot arm carries the pen to a preset position and orientation before the next spin action is executed (Fig~\ref{fig:task} Grasp and Pre-spin Pose). This process consistently resets the system. The trajectories are captured using an RGB-Depth camera in front of the robot arm. The camera has setup to have its $z$-axis pointing roughly towards the $m3$ finger when the MOE hand reaches the pre-spin configuration. Fig.~\ref{fig:setup} shows the setup.

\subsection{Pen Spinning Action Parameterization} 
We formulate the pen spinning task to be composed of a grasping action, a spinning action, and a catching action. Instead of optimizing for all DOFs in the system, we parameterize the pen spinning actions into a reduced set of variables: 
\textbf{Servo targets}: These are the target angle for each of the servo's internal PD controllers to reach to spin the pen. With three fingers and two servos per finger, this leads to a total of 6 parameters. We denote this component as $s \in \mathbb{R}^6$. Instead of using an absolute servo target, we choose to let $s$ represent servo angle changes with respect to the current servo angles. \textbf{Delay time}: Inspired by human pen-spinning, we observed that only the finger $m1$ is required to bend inward to catch the spinning pen, while fingers $m2$, $m3$ can remain stationary. Therefore, we do not search for the servo target angles for the catching action. Servos on finger $m1$ move to the inverse angles used during the spinning action, that is, if the two servos on $m1$ executed $\theta_1, \theta_2$ during spinning, they will execute $-\theta_1, -\theta_2$ during catching. Depending on the spinning action, the angular velocity of the pen will be different, leading to a different amount of time that the finger $m1$ needs to stay extended to not block the spinning pen. Therefore we still include one searchable parameter for this delay between the end of the spinning action and the beginning of the catching action. We denote this parameter as $d\in \mathbb{R}$. \textbf{Grasping location}: We add a searchable parameter to control the grasp location before spinning. This single parameter controls the  displacement from the grasping position to the center of the pen. We denote this parameter as $g \in \mathbb{R}$. The robot arm is pre-programmed to move the fingers to the center of the pen, and then adjust the end-effector position horizontally according to the grasping location parameter. The MOE fingers close according to a fixed sequence of motion to grasp the pen according to the grasping position.
In evaluation, we denote the action parameterization that contains spinning action servo targets and delay time as $(s,d) \in \mathbb{R}^7$. We denote the action parameterization that contains all the three components above as $(s,d,g)\in \mathbb{R}^8$

\subsection{State estimation and Optimization objective}

\begin{figure}[t]
  \centering
    \includegraphics[width=1.0\linewidth]{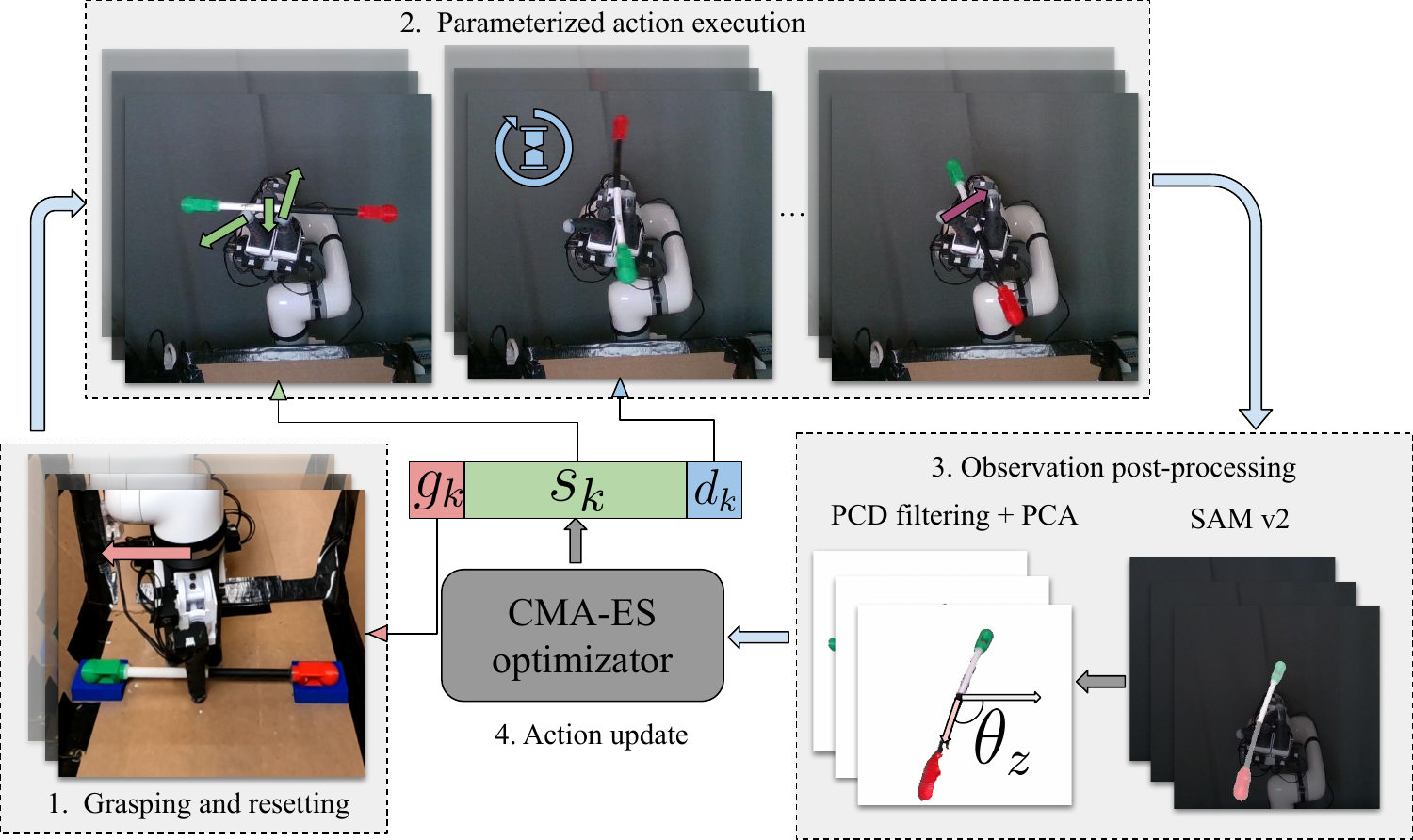}
    \caption{\algname{} optimization pipeline. There are 4 main stages for each iteration $k$: 1) During grasping and resetting, the robot arm moves the MOE hand to a target grasp location following a specific grasping location $g_k$. 2) The robot arm then moves the MOE hand to the pre-spin configuration, where the MOE fingers execute the parameterized action. 3) An RGB-D camera records the trial, and we apply masks from SAM-v2 to create a segmented point cloud. We then apply other post-processing of the point cloud to get the rotation and displacement state of the pen. 4) Lastly, the pipeline evaluates the objective function with observed states of the pen and updates the action parameters with the optimization algorithm CMA-ES. }
    \label{fig:pipeline}
\end{figure}

Fig.~\ref{fig:pipeline} summarizes the full \algname{} pipeline. To compute a reward, the system observes the state of the pen using RGB images and a point cloud captured by the RGB-D camera at 30\,fps. On the first frame of each trajectory, the system uses the Hough circle transform to locate red and green spherical markers on the pen. Segment Anything v2 \cite{ravi2024sam} then uses the pixel coordinates of the centers of the spheres as initial key points to create dense segmentation mask on each frame of the pen along its trajectory. The segmentation masks help select 3D points belonging to the pen. A bounding box around the MOE fingertips then filters out outlier points from the segmented point cloud and also indicates whether the pen is near the fingers. We consider the pen to be dropped in a frame if the filtered point cloud contains less than a threshold number of points. 
To retrieve the rotation state of the pen, the system then applies PCA on the filtered point cloud. The orientation of the pen is represented by the direction of the first principal component. We chose PCA instead of directly using the depth information of the center of spherical markers to increase robustness against noisy RGB-D data. The system finally projects the first principal component vector onto the $x$, $y$, and $z$ planes to compute the Euler angles of the pen in the camera coordinate system.

The objective function contains a reward term and a fall penalty term. The system computes the objective at each frame $t$ in a trajectory with $T$ total frames. The rotation reward is \[r_{\mathrm{rot}} = \frac{\Sigma_{t=0}^{T} \mathbf{1}^{\lVert p_t \rVert > n}(\theta_{z}^{t}-\theta_{z}^{t-1})}{2\pi},\] where $\theta_z^{t}$ is the rotation angle of the length of the pen around the $z$-axis in the camera coordinate frame at time step $t$. The indicator function $\mathbf{1}^{\lVert p_t \rVert > n}$ evaluates to 1 if the number of filtered points on the pen in a frame $p_t$ is greater than a threshold $n$; it is 0 otherwise. The depth camera points its $z$-axis towards and parallel to finger $m3$, and thus this rotation reward encourages rotation of the pen around the finger $m3$. The penalty term, \[p_{\mathrm{fall}} = \frac{\Sigma_{t=0}^{T} \mathbf{1}^{\lVert p_t \rVert > n}}{T},\] penalizes frames where the pen is displaced too far away from the fingers according to the indicator function's threshold. We apply a weight factor $\lambda$ to combine both terms into the final objective function:
\begin{equation}
    r =  r_{\mathrm{rot}} - \lambda p_{\mathrm{fall}}.
    \label{eq:objective}
\end{equation}

\subsection{Self-Supervised Primitive Parameter Optimization}
\algname{} uses Covariance Matrix Adaptation Evolution Strategy (CMA-ES) \cite{542381} to optimize the action parameters. CMA-ES is a gradient-free evolution strategy suitable for optimizing nonconvex objective functions such as Eq.~\ref{eq:objective}. At each generation, CMA-ES samples a population of action parameters from a multivariate normal distribution, parameterized by a mean and covariance matrix which are updated using the best performing candidates in the current generation. 
To prevent the robot arm from moving to a grasping location off the pen and the MOE hand from executing actions beyond its mechanical constraints, we constrain the output of the optimization algorithm to always be within the allowable range for the variable.

\renewcommand{\baselinestretch}{0.979} 
\section{Evaluation}

\subsection{Experiment setup}
We setup an environment (Fig.~\ref{fig:setup}) for a repeatable pen grasp and camera observation. The setup also includes a cage to facilitate resets by a human.
We test on three pen configurations to evaluate \algname{}'s robustness to varying physical properties of the object. 

\begin{figure*}[t]
  \centering
    \includegraphics[width=1.0\linewidth]{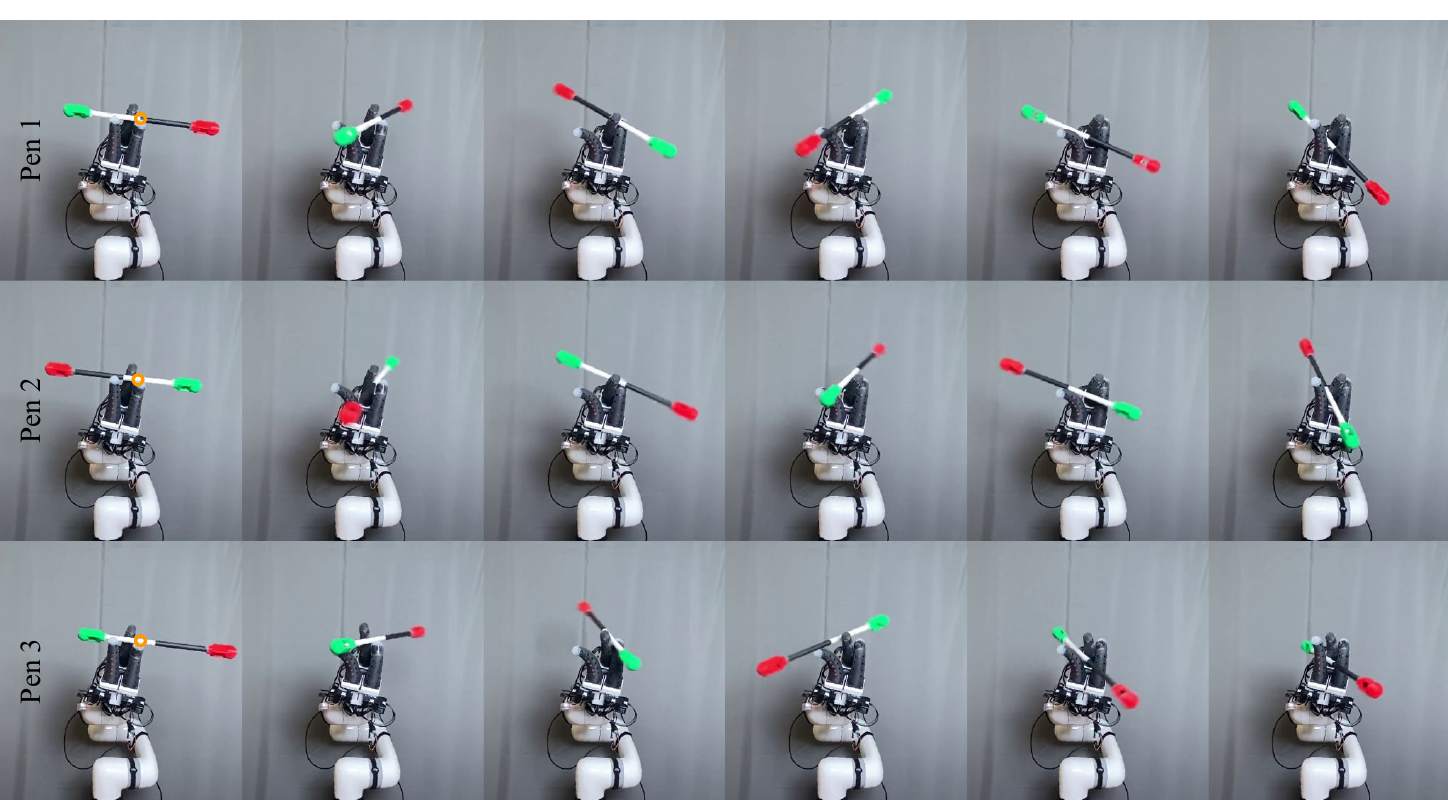}
    \caption{\textbf{Spinning visualization after optimization}. Top row: pen 1 with balanced weights. Middle row: pen 2 with unbalanced weight. Bottom row: pen 3 with unbalanced weight. The circle in the initial frame indicates the center of mass for the pen.}
    \label{fig:results}
\end{figure*}

\begin{figure*}[t]
  \centering
    \includegraphics[width=1.0\linewidth]{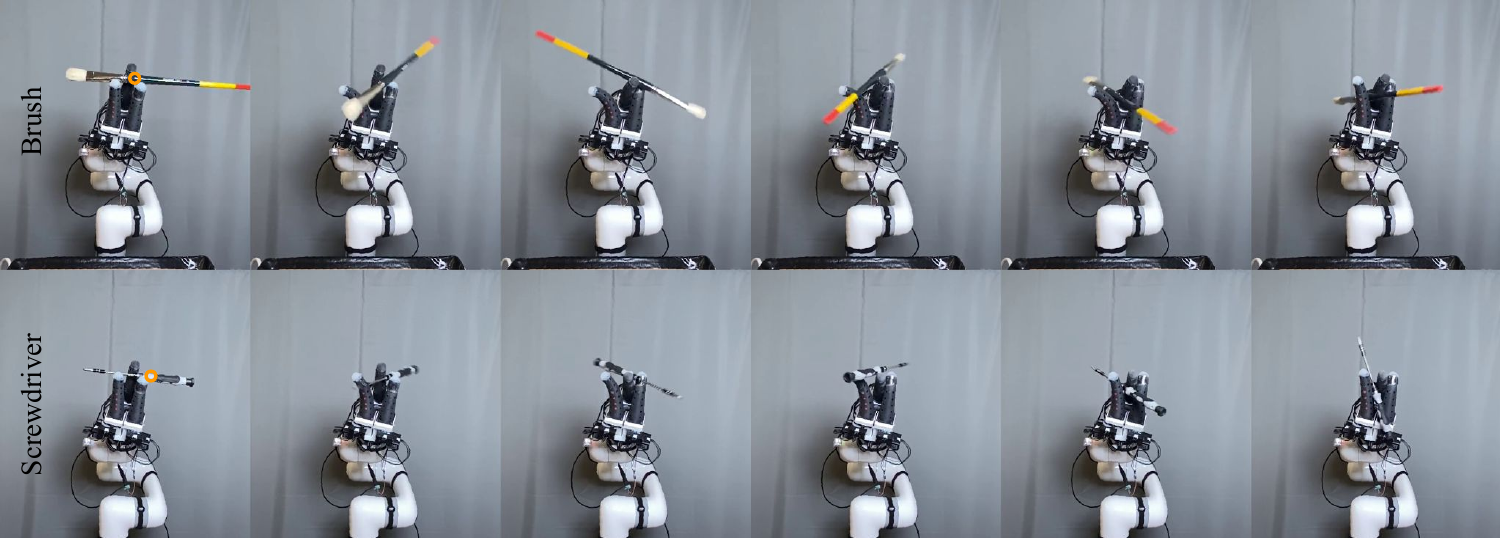}
    \caption{\textbf{Generalization to other objects}. We applied \algname{} to other objects with more irregular shapes, such as a brush or a screwdriver. The circle in the initial frame indicates the approximated center of masses. 
    }

    \label{fig:brush}
\end{figure*}

All three of the pens are 304\,mm long with a radius of 4.25\,mm and are visually identical. Pen 1 is a balanced pen where the center of mass is directly at the center of the length of the pen with a total mass of 38\,g. Pen 2 has a mass of 26\,g and is weighted so that the center of mass is offset from the center. This is achieved by removing a detachable weight object near the red spherical marker. Pen 3 is flipped to have the center of mass toward the other side. The screwdriver weighs 38\,g and has a length of 216\,mm. The maximum radius of the screwdriver is 20.5\,mm and the minimum radius is 3.8\,mm. The brush weighs 42\,g and has a length of 352\,mm. The maximum thickness of the brush is 21.3\,mm, and the minimum thickness is 6.7\,mm. We optimize the action parameters over 10 generations and evaluate the repeatability of the action parameters over 10 trials with the pen. Following the heuristics from the Hansen and Ostermeier~\cite{542381}, we choose the population size of CMA-ES to be $4 + 3 \log_2{8} \approx 13$ since our action parameterization has at most 8 dimensions. For evaluation on the brush and screwdriver, we optimize until the end of the first generations where we start to observe successful spins in the population or terminate at 10 generations. We then chose the first manually observed success for evaluation, rather than directly using the stored value of CMA-ES. The number of generations are later observed to be 4 when the first successful spins begin to be sampled for both the brush and the screwdriver.





\begin{table}[t]
\centering
\caption{\textsc{Action parameterization success rate} We optimized various action parameterizations using 10 generations of \algname{}.
The results suggest that optimizing both grasp location and spinning parameters yields the best performance, with generalization demonstrated on non-pen objects with varying geometries and mass distributions.}
\begin{tabular}{@{} lc cc@{}}
\toprule
    Action Parameterization & Parameters & Object & Successes \\
\midrule
     \multirow{3}{*}{Initialization} & \multirow{3}{*}{$\emptyset
$}  & pen 1 & \phantom10 / 10 \\
     & & pen 2 & \phantom10 / 10 \\
     & & pen 3 & \phantom10 / 10 \\
     \cmidrule{1-4} 
     
     \multirow{3}{*}{No grasp optimization} & \multirow{3}{*}{$(s,d)$}  & pen 1 & \phantom10 / 10 \\
     & & pen 2 & \phantom17 / 10 \\
     & & pen 3 & \phantom10 / 10 \\
     \cmidrule{1-4} 

     \multirow{3}{*}{Optimal action from Pen 1} & \multirow{3}{*}{$(s,d,g)$}  & pen 1 & \bf 10 / 10 \\
     & & pen 2 & \phantom10 / 10 \\
     & & pen 3 & \phantom17 / 10 \\
     \cmidrule{1-4} 
     
     \multirow{5}{*}{Full optimization (proposed)} & \multirow{5}{*}{$(s,d,g)$}  & pen 1 & \bf 10 / 10 \\
     & & pen 2 & \bf 10 / 10 \\
     & & pen 3 & \bf 10 / 10 \\
     & & brush & 10 / 10 \\
     & & screwdriver & \phantom15 / 10 \\
\bottomrule
\end{tabular}
\label{tab:ablation} 
\end{table}



\subsection{Results}
Fig.~\ref{fig:results} shows successful pen spins after optimization. During optimization, 
the reward function we used only indirectly captures whether a spinning action is successful or not. Thus, a human observer labels trials a success or failure. A trail is a success if the pen spins over finger $m3$ and does not fall off the fingers.
Table~\ref{tab:ablation} reports the success rates of each baseline and ablated method. We initialized the CMA-ES optimization with heuristically hand-crafted action parameters. However, directly applying a fixed action initialization does not lead to any success in all three pen settings, each failing with 0/10 success rates (row 1 in the table). The result indicates to us that optimizing the actions for the MOE hand specifically is important for the success in these tasks. 

We compared \algname{} optimization with all of the action parameters against different action parameterizations and report the results in Table~\ref{tab:ablation}. In the \emph{Initialization} row, \algname{} does not optimize the grasp action $g$ and always grasps the center of the pen. 
In the \emph{No grasp optimization}, we optimize $(s,d)$, but not the grasp point $g$. The robot again always grasps the pen center.
With this experiment, we see 
the efficacy of optimizing only $(s,d)$ is highly object-dependent. With a central grasp, we could only succeed for pen 2 with a success rate of 7/10. A reason for grasping the center of the pen's length working for pen 2 could be that the optimal grasping point for pen 2 is the closest to the center of the pen as we can see in Fig.~\ref{fig:results}. These results highlight the importance of optimizing for the grasping point and the spinning action parameters for the system to work well for varying pen properties.
In the \emph{Optimal action from Pen 1} row, we show the results of 
optimizing action parameters for pen 1, then applying the action to pens 2 and 3. This shows the necessity to update action parameters for each new object. 
Directly using the optimal $(s,d,g)$ parameters from pen 1 to pen 3 results in 7/10 successes, while the same set of parameters results in 0/10 successes on pen 2. 
We can see in Fig.~\ref{fig:results} that optimal grasping points between pen 1 and pen 3 are both left of the pen's center in the image frame. This may explain why the optimal actions from pen 1 had some success on pen 3. In these experiments, all three pens are visually identical and therefore depend on \algname{}'s ability to interact with the object to search for optimal action parameters. 
In the \emph{Full optimization (proposed)} row, we found that optimizing $(s,d,g)$ for each object results in 10/10 success rates for all pens. The higher success rate for pen 2 using full optimization compared to not optimizing grasping also suggests that having the ability to search over the grasping position enabled the search for a more robost spinning motion. 

Lastly, we experiment with \algname{} applied to two other objects: a brush and a screwdriver. Fig~\ref{fig:brush} shows the results of these generalization experiments. \algname{} achieves 10/10 and 5/10 success rates for the brush and screwdriver respectively. The screwdriver is particularly challenging to spin because of its irregular shape. However, \algname{} optimized the action parameters achieves successful spins, highlighting \algname{}'s versatility.


\renewcommand{\baselinestretch}{0.979} 

\section{Conclusion}
In this work, we present \algname{}, a robust system for dynamic in-hand pen spinning with a soft robotic end-effector. \algname{} leverages real-world interactions to learn from trial-and-error to optimize the pen grasping and spinning actions for soft robotic end-effectors. Importantly, it does not require explicit knowledge of the object's physical properties, allowing the system to robustly spin pens even when the weight distributions and shapes are varied. By using a sampling-based optimization strategy, we were able to efficiently explore the action space and discover the optimal set of actions for pen spinning. 

We demonstrated the system's robustness across pens of different weights and  weight distributions, suggesting the ability to generalize and adapt to changes in object inertial properties that are not easily observable without interaction. Additionally, we tested \algname{} on two other objects,  a brush and a screwdriver, and found that \algname{} still succeeded in optimizing action parameters to spin them. The results highlighted the effectiveness of soft robotic end-effectors performing dynamic manipulation tasks with contact-rich and dynamic conditions.

 In the future, we will focus on extending and generalizing the approach to objects beyond pen-shaped objects and explore other in-hand dynamic tasks with soft robotic end-effectors. We also hope to improve the system's efficiency and reliability by incorporating a wider range of sensory feedback such as proprioception and contact estimation to enhance the system's performance and generalizability. 



\addtolength{\textheight}{-0.2cm}   


{\footnotesize
\bibliographystyle{ieeetr}
\bibliography{references.bib}
}

\newpage

\end{document}